\documentclass{article} % For LaTeX2e
\usepackage{nips13submit_e,times}
\usepackage{hyperref}
\usepackage{url}
\usepackage{amsmath}
\usepackage{graphicx}
\usepackage{caption}
\usepackage{subcaption}
\usepackage{multirow}

\title{Network In Network}

\author{%
Min Lin\textsuperscript{1,2}, Qiang Chen\textsuperscript{2}, Shuicheng
Yan\textsuperscript{2}\\
\textsuperscript{1}Graduate School for Integrative Sciences and
Engineering\\
\textsuperscript{2}Department of Electronic \& Computer Engineering\\
National University of Singapore, Singapore\\
\texttt{\{linmin,chenqiang,eleyans\}@nus.edu.sg}
}

\nipsfinalcopy % Uncomment for camera-ready version

\begin{document}

\maketitle

\begin{abstract}
  We propose a novel deep network structure called ``\textbf{N}etwork
  \textbf{I}n \textbf{N}etwork''(NIN) to enhance model
  discriminability for local patches within the receptive field. The
  conventional convolutional layer uses linear filters followed by a
  nonlinear activation function to scan the input. Instead, we build
  micro neural networks with more complex structures to abstract the
  data within the receptive field. We instantiate the micro neural
  network with a multilayer perceptron, which is a potent function
  approximator. The feature maps are obtained by sliding the micro
  networks over the input in a similar manner as CNN; they are then
  fed into the next layer. Deep NIN can be implemented by stacking
  mutiple of the above described structure. With enhanced local
  modeling via the micro network, we are able to utilize global
  average pooling over feature maps in the classification layer, which
  is easier to interpret and less prone to overfitting than traditional
  fully connected layers. We demonstrated the state-of-the-art
  classification performances with NIN on CIFAR-10 and CIFAR-100, and
  reasonable performances on SVHN and MNIST datasets.
\end{abstract}

\section{Introduction}
Convolutional neural networks (CNNs) \cite{lecun1998gradient} consist of
alternating convolutional layers and pooling layers. Convolution
layers take inner product of the linear filter and the underlying
receptive field followed by a nonlinear activation function at every
local portion of the input. The resulting outputs are called feature
maps.

The convolution filter in CNN is a generalized linear model (GLM) for
the underlying data patch, and we argue that the level of abstraction
is low with GLM. By abstraction we mean that the feature is invariant
to the variants of the same
concept \cite{bengio2013representation}. Replacing the GLM with a more
potent nonlinear function approximator can enhance the abstraction
ability of the local model. GLM can achieve a good extent of
abstraction when the samples of the latent concepts are linearly
separable, i.e. the variants of the concepts all live on one side of
the separation plane defined by the GLM. Thus conventional CNN
implicitly makes the assumption that the latent concepts are linearly
separable. However, the data for the same concept often live on a
nonlinear manifold, therefore the representations that capture these
concepts are generally highly nonlinear function of the input. In NIN,
the GLM is replaced with a "micro network" structure which is a
general nonlinear function approximator. In this work, we choose
multilayer perceptron \cite{rosenblatt1961principles} as the
instantiation of the micro network, which is a universal function
approximator and a neural network trainable by back-propagation.

\begin{figure}
  \begin{subfigure}{.5\textwidth}
    \centering
    \includegraphics[height=.5\linewidth]{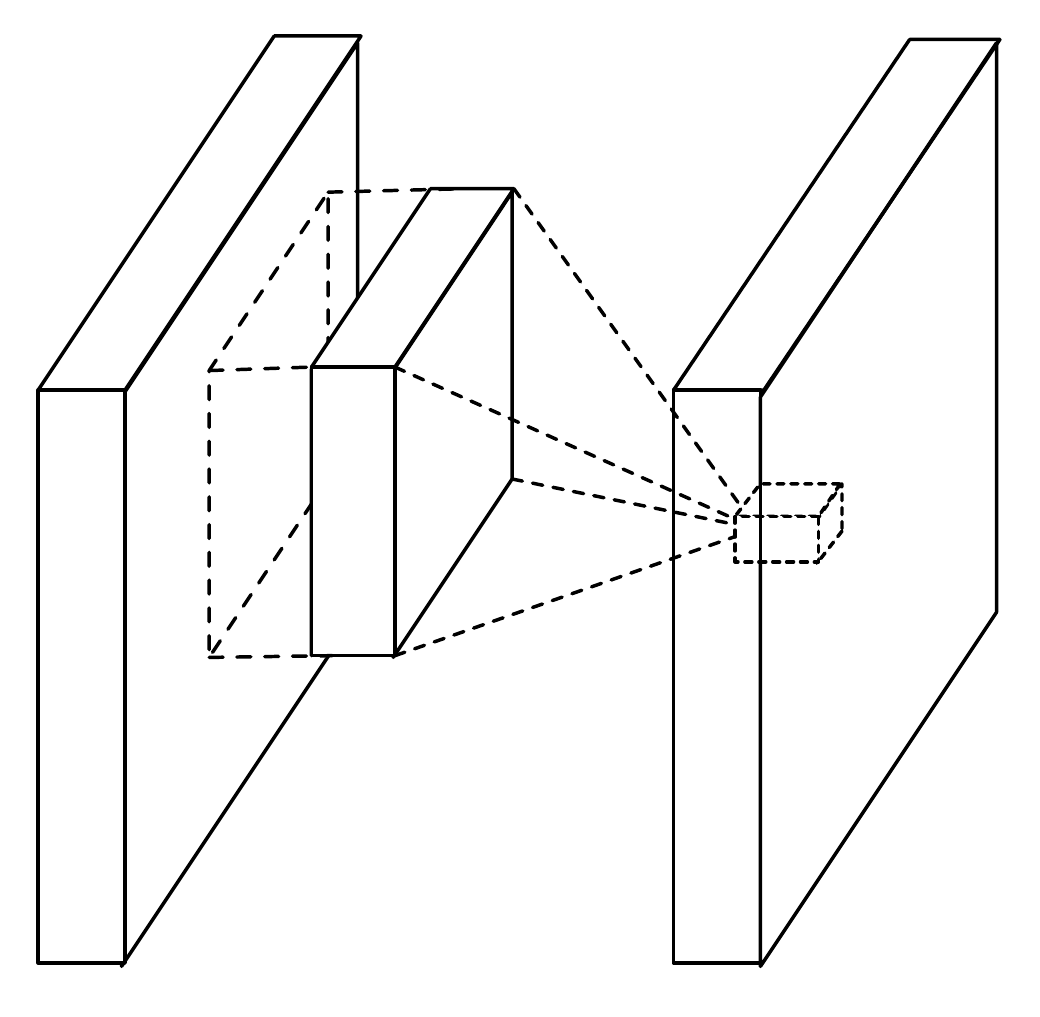}
    \caption{Linear convolution layer}
    \label{fig:linearconv}
  \end{subfigure}
  \begin{subfigure}{.5\textwidth}
    \centering
    \includegraphics[height=.5\linewidth]{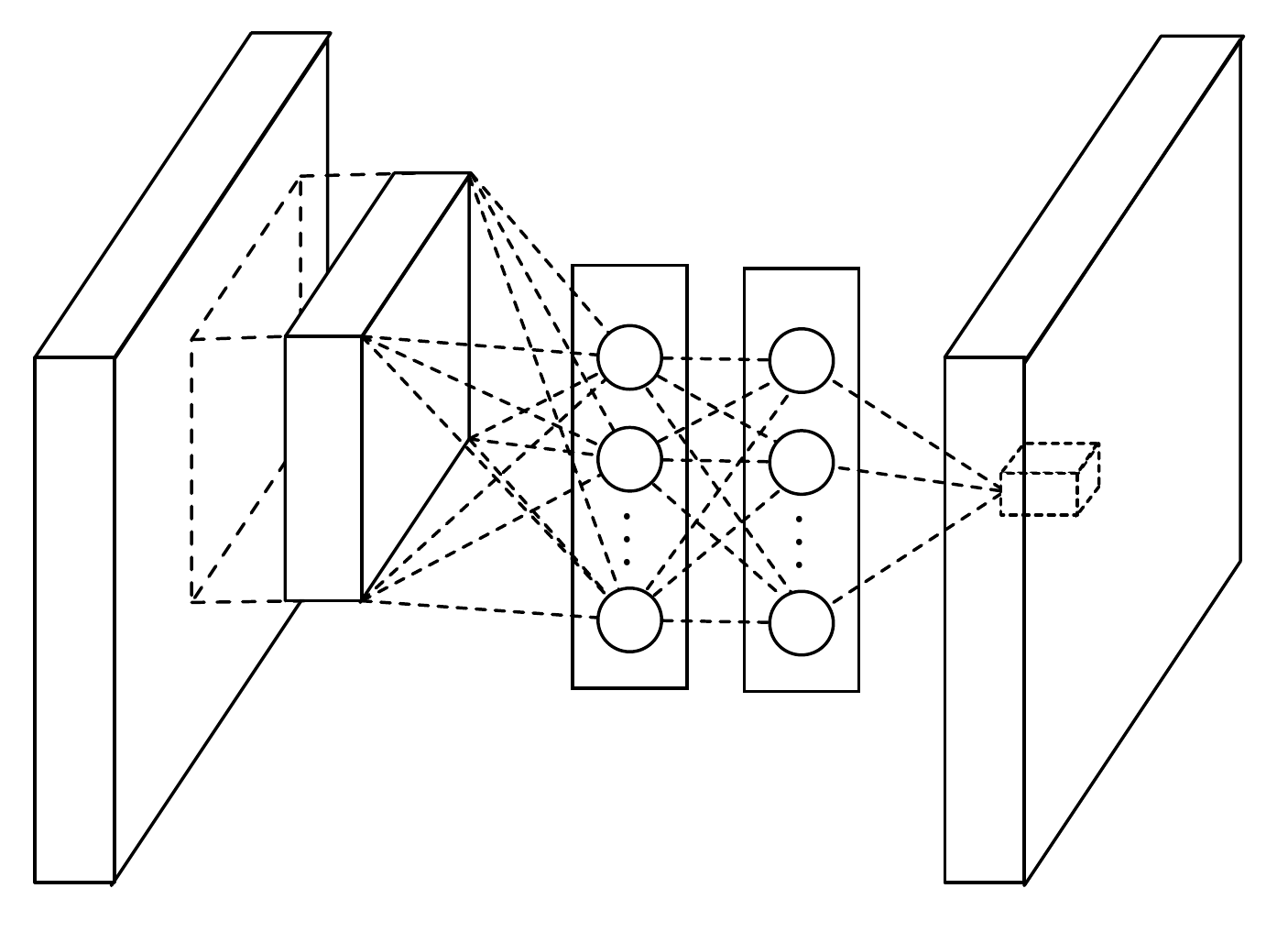}
    \caption{Mlpconv layer}
    \label{fig:mlpconv}
  \end{subfigure}
  \caption{Comparison of linear convolution layer and mlpconv
    layer. The linear convolution layer includes a linear filter while
  the mlpconv layer includes a micro network (we choose the
  multilayer perceptron in this paper). Both layers map the local
  receptive field to a confidence value of the latent concept.}\label{fig:compareconv}
\end{figure}

The resulting structure which we call an mlpconv layer is compared
with CNN in Figure \ref{fig:compareconv}. Both the linear
convolutional layer and the mlpconv layer map the local receptive
field to an output feature vector. The mlpconv maps the input local
patch to the output feature vector with a multilayer perceptron (MLP)
consisting of multiple fully connected layers with nonlinear
activation functions. The MLP is shared among all local receptive
fields. The feature maps are obtained by sliding the MLP over the
input in a similar manner as CNN and are then fed into the next
layer. The overall structure of the NIN is the stacking of multiple
mlpconv layers. It is called ``Network In Network'' (NIN) as we have
micro networks (MLP), which are composing elements of the overall deep
network, within mlpconv layers,

Instead of adopting the traditional fully connected layers for
classification in CNN, we directly output the spatial average of the
feature maps from the last mlpconv layer as the confidence of
categories via a global average pooling layer, and then the resulting
vector is fed into the softmax layer. In traditional CNN, it is
difficult to interpret how the category level information from the
objective cost layer is passed back to the previous convolution layer
due to the fully connected layers which act as a black box in
between. In contrast, global average pooling is more meaningful and
interpretable as it enforces correspondance between feature maps and
categories, which is made possible by a stronger local modeling using
the micro network. Furthermore, the fully connected layers are prone
to overfitting and heavily depend on dropout regularization
\cite{krizhevsky2012imagenet} \cite{hinton2012improving}, while global
average pooling is itself a structural regularizer, which natively
prevents overfitting for the overall structure.

\section{Convolutional Neural Networks}

Classic convolutional neuron networks \cite{lecun1998gradient} consist
of alternatively stacked convolutional layers and spatial pooling
layers. The convolutional layers generate feature maps by linear
convolutional filters followed by nonlinear activation functions
(rectifier, sigmoid, tanh, etc.).  Using the linear rectifier as an
example, the feature map can be calculated as follows:

\begin{equation}
  f_{i,j,k}=\text{max}(w_k^Tx_{i,j}, 0).
  \label{equ:linrelu}
\end{equation}

Here $(i,j)$ is the pixel index in the feature map, $x_{ij}$ stands
for the input patch centered at location $(i,j)$, and $k$ is used to
index the channels of the feature map.

This linear convolution is sufficient for abstraction when the
instances of the latent concepts are linearly separable. However,
representations that achieve good abstraction are generally highly
nonlinear functions of the input data. In conventional CNN, this might
be compensated by utilizing an over-complete set of filters
\cite{le2011ica} to cover all variations of the latent
concepts. Namely, individual linear filters can be learned to detect
different variations of a same concept. However, having too many
filters for a single concept imposes extra burden on the next layer,
which needs to consider all combinations of variations from the
previous layer \cite{goodfellow2013piecewise}. As in CNN, filters from
higher layers map to larger regions in the original input. It
generates a higher level concept by combining the lower level concepts
from the layer below. Therefore, we argue that it would be beneficial
to do a better abstraction on each local patch, before combining
them into higher level concepts.

In the recent maxout network \cite{goodfellow2013maxout}, the number
of feature maps is reduced by maximum pooling over affine feature maps
(affine feature maps are the direct results from linear convolution
without applying the activation function). Maximization over linear
functions makes a piecewise linear approximator which is capable of
approximating any convex functions. Compared to conventional
convolutional layers which perform linear separation, the maxout network
is more potent as it can separate concepts that lie within convex
sets. This improvement endows the maxout network with the best
performances on several benchmark datasets.

However, maxout network imposes the prior that instances of a latent
concept lie within a convex set in the input space, which does not
necessarily hold. It would be necessary to employ a more general
function approximator when the distributions of the latent concepts
are more complex. We seek to achieve this by introducing the novel
``Network In Network'' structure, in which a micro network is
introduced within each convolutional layer to compute more abstract
features for local patches.

Sliding a micro network over the input has been proposed in several
previous works. For example, the Structured Multilayer Perceptron
(SMLP) \cite{gulccehre2013knowledge} applies a shared multilayer
perceptron on different patches of the input image; in another work, a
neural network based filter is trained for face detection
\cite{rowley1995human}. However, they are both designed for specific
problems and both contain only one layer of the sliding network
structure. NIN is proposed from a more general perspective, the micro
network is integrated into CNN structure in persuit of better
abstractions for all levels of features.

\section{Network In Network}
We first highlight the key components of our proposed ``Network In
Network'' structure: the MLP convolutional layer and the global
averaging pooling layer in Sec. 3.1 and Sec. 3.2 respectively. Then we
detail the overall NIN in Sec. 3.3.

\subsection{MLP Convolution Layers}

Given no priors about the distributions of the latent concepts, it is
desirable to use a universal function approximator for feature
extraction of the local patches, as it is capable of approximating
more abstract representations of the latent concepts. Radial basis
network and multilayer perceptron are two well known universal
function approximators. We choose multilayer perceptron in this work
for two reasons. First, multilayer perceptron is compatible with the
structure of convolutional neural networks, which is trained using
back-propagation. Second, multilayer perceptron can be a deep model
itself, which is consistent with the spirit of feature re-use
\cite{bengio2013representation}.  This new type of layer is called
mlpconv in this paper, in which MLP replaces the GLM to convolve over
the input. Figure \ref{fig:compareconv} illustrates the difference
between linear convolutional layer and mlpconv layer. The calculation
performed by mlpconv layer is shown as follows:

\begin{eqnarray}
  \label{mlpconvfeaturemapcalc}
  f_{i,j,k_1}^1&=&\text{max}({w_{k_1}^1}^Tx_{i,j}+b_{k_1},0). \nonumber \\
  &\vdots{}& \nonumber \\
  f_{i,j,k_n}^n&=&\text{max}({w_{k_n}^n}^Tf_{i,j}^{n-1}+b_{k_n},0).
\end{eqnarray}

Here $n$ is the number of layers in the multilayer
perceptron. Rectified linear unit is used as the activation function
in the multilayer perceptron.

From cross channel (cross feature map) pooling point of view, Equation
\ref{mlpconvfeaturemapcalc} is equivalent to cascaded cross channel
parametric pooling on a normal convolution layer. Each pooling layer
performs weighted linear recombination on the input feature maps,
which then go through a rectifier linear unit. The cross channel
pooled feature maps are cross channel pooled again and again in the
next layers. This cascaded cross channel parameteric pooling structure
allows complex and learnable interactions of cross channel
information.

The cross channel parametric pooling layer is also equivalent to a
convolution layer with 1x1 convolution kernel. This interpretation
makes it straightforawrd to understand the structure of NIN.

\textbf{Comparison to maxout layers:} the maxout layers in the maxout
network performs max pooling across multiple affine feature maps
\cite{goodfellow2013maxout}. The feature maps of maxout layers are
calculated as follows:

\begin{equation}
  \label{equ:maxout}
  f_{i,j,k}=\max_m{}(w_{k_m}^Tx_{i,j}).
\end{equation}

Maxout over linear functions forms a piecewise linear function which
is capable of modeling any convex function. For a convex function,
samples with function values below a specific threshold form a convex
set. Therefore, by approximating convex functions of the local patch,
maxout has the capability of forming separation hyperplanes for
concepts whose samples are within a convex set (i.e. $l_2$ balls,
convex cones). Mlpconv layer differs from maxout layer in that the
convex function approximator is replaced by a universal function
approximator, which has greater capability in modeling various
distributions of latent concepts.

\subsection{Global Average Pooling}

Conventional convolutional neural networks perform convolution in the
lower layers of the network. For classification, the feature maps of
the last convolutional layer are vectorized and fed into fully
connected layers followed by a softmax logistic regression layer
\cite{krizhevsky2012imagenet} \cite{goodfellow2013maxout}
\cite{zeiler2013stochastic}. This structure bridges the convolutional
structure with traditional neural network classifiers. It treats the
convolutional layers as feature extractors, and the resulting feature
is classified in a traditional way.

However, the fully connected layers are prone to overfitting, thus
hampering the generalization ability of the overall network.  Dropout
is proposed by Hinton et al. \cite{hinton2012improving} as a
regularizer which randomly sets half of the activations to the fully
connected layers to zero during training. It has improved the
generalization ability and largely prevents overfitting
\cite{krizhevsky2012imagenet}.

In this paper, we propose another strategy called global average
pooling to replace the traditional fully connected layers in CNN. The
idea is to generate one feature map for each corresponding category of
the classification task in the last mlpconv layer. Instead of adding
fully connected layers on top of the feature maps, we take the average
of each feature map, and the resulting vector is fed directly into the
softmax layer. One advantage of global
average pooling over the fully connected layers is that it is more
native to the convolution structure by enforcing correspondences
between feature maps and categories. Thus the feature maps can be
easily interpreted as categories confidence maps. Another advantage is
that there is no parameter to optimize in the global average pooling
thus overfitting is avoided at this layer. Futhermore, global average
pooling sums out the spatial information, thus it is more robust to
spatial translations of the input.

We can see global average pooling as a structural regularizer that
explicitly enforces feature maps to be confidence maps of concepts
(categories). This is made possible by the mlpconv layers, as they
makes better approximation to the confidence maps than GLMs.

\subsection{Network In Network Structure}

\begin{figure}
  \centering
  \includegraphics[width=.9\linewidth]{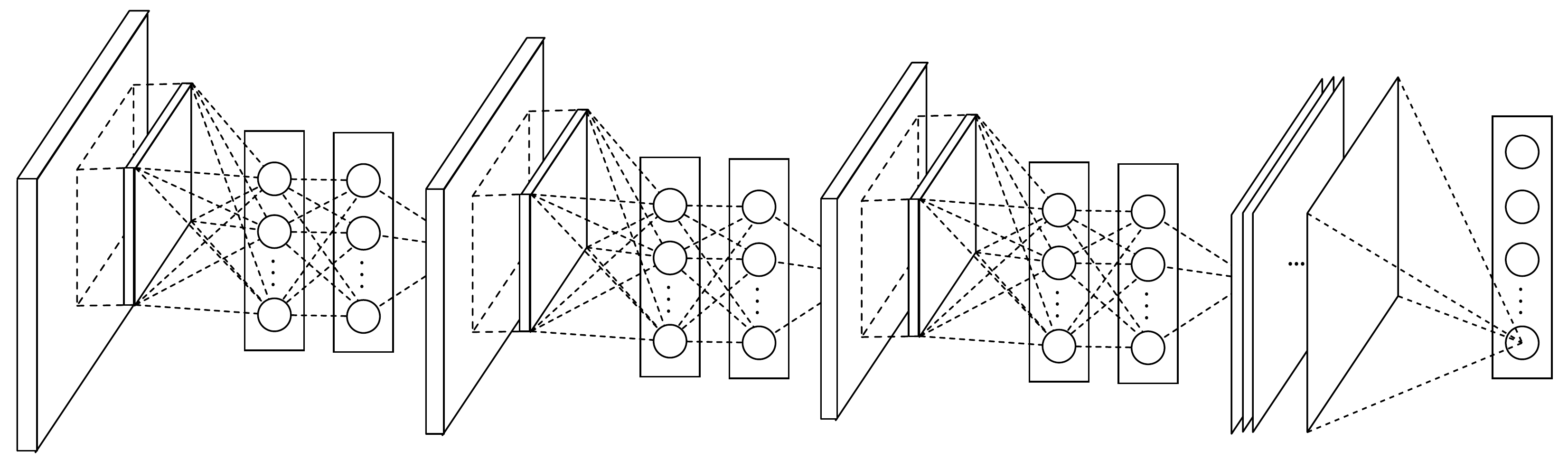}
  \caption{The overall structure of Network In Network. In this paper
    the NINs include the stacking of three mlpconv layers and one
    global average pooling layer.}
  \label{fig:NIN}
\end{figure}

The overall structure of NIN is a stack of mlpconv layers, on top of
which lie the global average pooling and the objective cost
layer. Sub-sampling layers can be added in between the mlpconv layers
as in CNN and maxout networks. Figure \ref{fig:NIN} shows an NIN with
three mlpconv layers. Within each mlpconv layer, there is a
three-layer perceptron. The number of layers in both NIN and the
micro networks is flexible and can be tuned for specific tasks.

\section{Experiments}

\subsection{Overview}
We evaluate NIN on four benchmark datasets: CIFAR-10
\cite{krizhevsky2009learning}, CIFAR-100
\cite{krizhevsky2009learning}, SVHN \cite{netzer2011reading} and MNIST
\cite{lecun1998gradient}. The networks used for the datasets all
consist of three stacked mlpconv layers, and the mlpconv layers in all
the experiments are followed by a spatial max pooling layer which
down-samples the input image by a factor of two. As a regularizer,
dropout is applied on the outputs of all but the last mlpconv
layers. Unless stated specifically, all the networks used in the
experiment section use global average pooling instead of fully
connected layers at the top of the network. Another regularizer
applied is weight decay as used by Krizhevsky et
al. \cite{krizhevsky2012imagenet}. Figure \ref{fig:NIN} illustrates
the overall structure of NIN network used in this section. The
detailed settings of the parameters are provided in the supplementary
materials. We implement our network on the super fast cuda-convnet
code developed by Alex Krizhevsky
\cite{krizhevsky2012imagenet}. Preprocessing of the datasets,
splitting of training and validation sets all follow Goodfellow et
al. \cite{goodfellow2013maxout}.

We adopt the training procedure used by Krizhevsky et
al. \cite{krizhevsky2012imagenet}. Namely, we manually set proper
initializations for the weights and the learning rates. The network is
trained using mini-batches of size 128. The training process starts
from the initial weights and learning rates, and it continues until the
accuracy on the training set stops improving, and then the learning rate
is lowered by a scale of 10. This procedure is repeated once such that
the final learning rate is one percent of the initial value.

\subsection{CIFAR-10}
The CIFAR-10 dataset \cite{krizhevsky2009learning} is composed of 10
classes of natural images with 50,000 training images in total, and
10,000 testing images. Each image is an RGB image of size 32x32. For
this dataset, we apply the same global contrast normalization and ZCA
whitening as was used by Goodfellow et al. in the maxout network
\cite{goodfellow2013maxout}. We use the last 10,000 images of the
training set as validation data.

The number of feature maps for each mlpconv layer in this experiment
is set to the same number as in the corresponding maxout network. Two
hyper-parameters are tuned using the validation set, i.e. the local
receptive field size and the weight decay. After that the
hyper-parameters are fixed and we re-train the network from scratch
with both the training set and the validation set. The resulting model
is used for testing.  We obtain a test error of 10.41\% on this
dataset, which improves more than one percent compared to the
state-of-the-art. A comparison with previous methods is shown in Table
\ref{cifar10}.

\newcommand{\specialcell}[2][c]{%
  \begin{tabular}[#1]{@{}c@{}}#2\end{tabular}}

\begin{table}[h!]
  \caption{Test set error rates for CIFAR-10 of various methods.}
  \label{cifar10}
  \begin{center}
    \begin{tabular}{lc}
      \hline
      Method         &Test Error \\
      \hline
      Stochastic Pooling \cite{zeiler2013stochastic}        &15.13\% \\
      CNN + Spearmint \cite{snoek2012practical}           &14.98\% \\
      Conv. maxout + Dropout \cite{goodfellow2013maxout}    &11.68\% \\
      \textbf{NIN + Dropout}  &\textbf{10.41\%} \\
      \hline
      CNN + Spearmint + Data Augmentation \cite{snoek2012practical} & 9.50\% \\
      Conv. maxout + Dropout + Data Augmentation
      \cite{goodfellow2013maxout} & 9.38\% \\
      DropConnect + 12 networks + Data Augmentation \cite{wan2013regularization} & 9.32\% \\
      \textbf{NIN + Dropout + Data Augmentation} &\textbf{8.81\%} \\
      \hline
    \end{tabular}
  \end{center}
\end{table}

\begin{figure}[h!]
  \centering
  \includegraphics[width=0.6\linewidth]{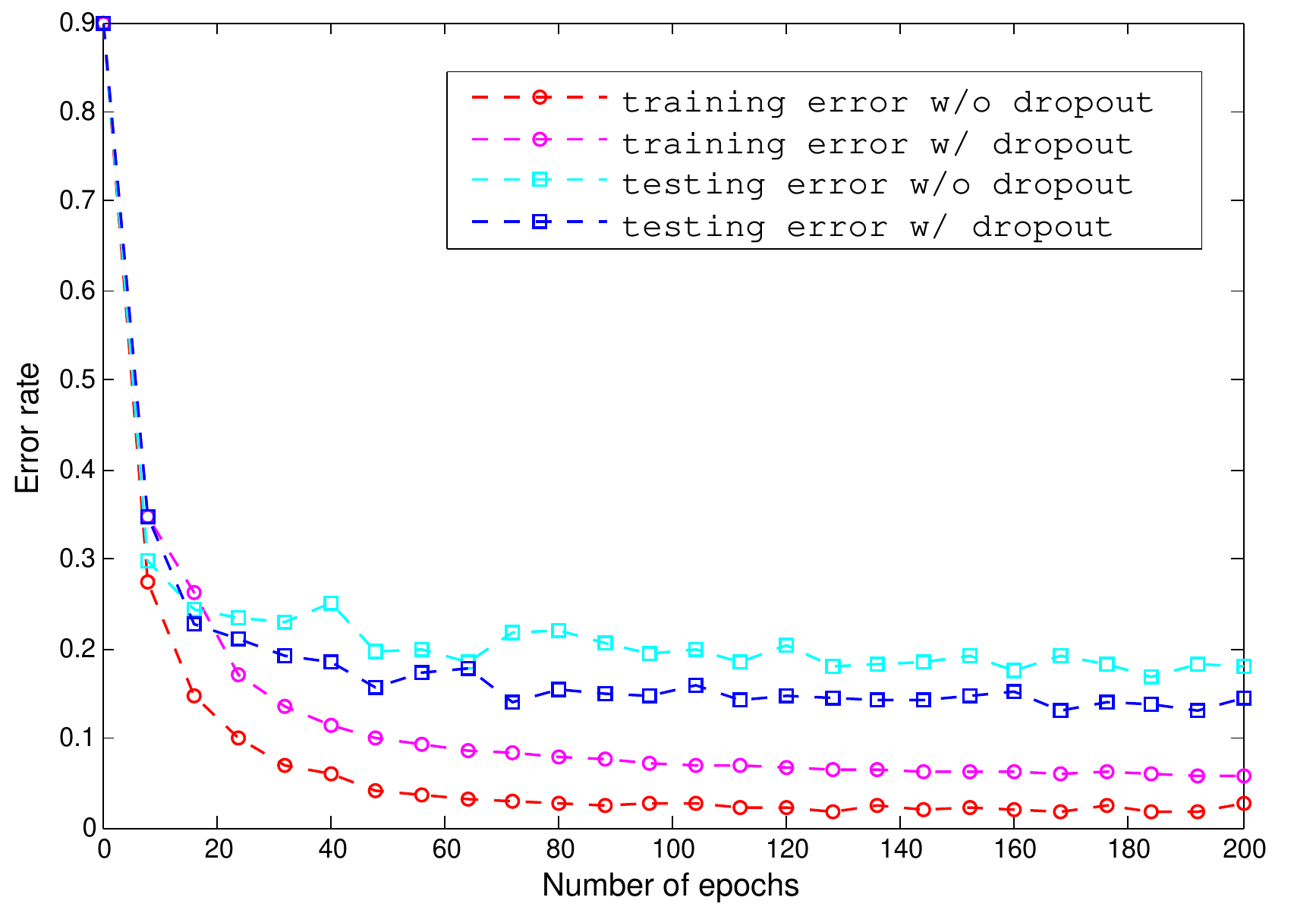}
  \caption{The regularization effect of dropout in between mlpconv
    layers. Training and testing error of NIN with and without dropout
  in the first 200 epochs of training is shown.}
  \label{fig:dropout}
\end{figure}

It turns out in our experiment that using dropout in between the
mlpconv layers in NIN boosts the performance of the network by
improving the generalization ability of the model. As is shown in
Figure \ref{fig:dropout}, introducing dropout layers in between the
mlpconv layers reduced the test error by more than 20\%. This
observation is consistant with Goodfellow et
al. \cite{goodfellow2013maxout}. Thus dropout is added in between the
mlpconv layers to all the models used in this paper. The model without
dropout regularizer achieves an error rate of 14.51\% for the CIFAR-10
dataset, which already surpasses many previous state-of-the-arts with
regularizer (except maxout). Since performance of maxout without
dropout is not available, only dropout regularized version are
compared in this paper.

To be consistent with previous works, we also evaluate our method on
the CIFAR-10 dataset with translation and horizontal flipping
augmentation. We are able to achieve a test error of 8.81\%, which
sets the new state-of-the-art performance.

\subsection{CIFAR-100}
The CIFAR-100 dataset \cite{krizhevsky2009learning} is the same in
size and format as the CIFAR-10 dataset, but it contains 100
classes. Thus the number of images in each class is only one tenth of
the CIFAR-10 dataset. For CIFAR-100 we do not tune the
hyper-parameters, but use the same setting as the CIFAR-10
dataset. The only difference is that the last mlpconv layer outputs
100 feature maps. A test error of 35.68\% is obtained for CIFAR-100
which surpasses the current best performance without data augmentation
by more than one percent. Details of the performance comparison are
shown in Table \ref{cifar100}.

\begin{table}[h!]
  \caption{Test set error rates for CIFAR-100 of various methods.}
  \label{cifar100}
  \begin{center}
    \begin{tabular}{lc}
      \hline
      Method         &Test Error \\
      \hline
      Learned Pooling \cite{malinowski2013learnable}        &43.71\% \\
      Stochastic Pooling \cite{zeiler2013stochastic}        &42.51\% \\
      Conv. maxout + Dropout \cite{goodfellow2013maxout}    &38.57\% \\
      Tree based priors \cite{srivastava2013discriminative} &36.85\% \\
      \textbf{NIN + Dropout}                       &\textbf{35.68\%} \\
      \hline
    \end{tabular}
  \end{center}
\end{table}

\subsection{Street View House Numbers}
The SVHN dataset \cite{netzer2011reading} is composed of 630,420 32x32
color images, divided into training set, testing set and an extra
set. The task of this data set is to classify the digit located at the
center of each image. The training and testing procedure follow
Goodfellow et al. \cite{goodfellow2013maxout}. Namely 400 samples per
class selected from the training set and 200 samples per class from
the extra set are used for validation. The remainder of the training
set and the extra set are used for training. The validation set is
only used as a guidance for hyper-parameter selection, but never used
for training the model.

\begin{table}[h!]
  \caption{Test set error rates for SVHN of various methods.}
  \label{svhn}
  \begin{center}
    \begin{tabular}{lc}
      \hline
      Method         &Test Error \\
      \hline
      Stochastic Pooling \cite{zeiler2013stochastic}        &2.80\% \\
      Rectifier + Dropout \cite{srivastava2013improving}        &2.78\% \\
      Rectifier + Dropout + Synthetic Translation \cite{srivastava2013improving} &2.68\%  \\
      Conv. maxout + Dropout \cite{goodfellow2013maxout}     &2.47\% \\
      NIN + Dropout  &2.35\% \\
      Multi-digit Number Recognition \cite{goodfellow2013multi}
      &2.16\% \\
      \textbf{DropConnect} \cite{wan2013regularization} &\textbf{1.94\%} \\
      \hline
    \end{tabular}
  \end{center}
\end{table}

Preprocessing of the dataset again follows Goodfellow et
al. \cite{goodfellow2013maxout}, which was a local contrast
normalization.  The structure and parameters used in SVHN are similar
to those used for CIFAR-10, which consist of three mlpconv layers
followed by global average pooling. For this dataset, we obtain a test
error rate of 2.35\%. We compare our result with methods that did not
augment the data, and the comparison is shown in Table \ref{svhn}.

\subsection{MNIST}
The MNIST \cite{lecun1998gradient} dataset consists of hand written
digits 0-9 which are 28x28 in size. There are 60,000 training images
and 10,000 testing images in total. For this dataset, the same network
structure as used for CIFAR-10 is adopted. But the numbers of feature
maps generated from each mlpconv layer are reduced. Because MNIST is a
simpler dataset compared with CIFAR-10; fewer parameters are
needed. We test our method on this dataset without data
augmentation. The result is compared with previous works that adopted
convolutional structures, and are shown in Table \ref{mnist}.

\begin{table}[h!]
  \caption{Test set error rates for MNIST of various methods.}
  \label{mnist}
  \begin{center}
    \begin{tabular}{lc}
      \hline
      Method         &Test Error \\
      \hline
      2-Layer CNN + 2-Layer NN \cite{zeiler2013stochastic}    &0.53\% \\
      Stochastic Pooling \cite{zeiler2013stochastic}          &0.47\% \\
      NIN + Dropout  &0.47\% \\
      \textbf{Conv. maxout + Dropout \cite{goodfellow2013maxout}} &\textbf{0.45\%} \\
      \hline
    \end{tabular}
  \end{center}
\end{table}

We achieve comparable but not better performance (0.47\%) than the
current best (0.45\%) since MNIST has been tuned to a very low error
rate.

\subsection{Global Average Pooling as a Regularizer}

Global average pooling layer is similar to the fully connected layer in
that they both perform linear transformations of the vectorized
feature maps. The difference lies in the transformation matrix. For
global average pooling, the transformation matrix is prefixed and it
is non-zero only on block diagonal elements which share the same
value. Fully connected layers can have dense transformation matrices
and the values are subject to back-propagation optimization. To study
the regularization effect of global average pooling, we replace the
global average pooling layer with a fully connected layer, while the
other parts of the model remain the same. We evaluated this model with
and without dropout before the fully connected linear layer. Both
models are tested on the CIFAR-10 dataset, and a comparison of the
performances is shown in Table \ref{gapfc}.

\begin{table}[h!]
  \caption{Global average pooling compared to fully connected layer.}
  \label{gapfc}
  \begin{center}
    \begin{tabular}{lc}
      \hline
      Method                   &Testing Error  \\
      \hline
      mlpconv + Fully Connected               &11.59\%  \\
      mlpconv + Fully Connected + Dropout     &10.88\%  \\
      mlpconv + Global Average Pooling        &10.41\%  \\
      \hline
      % linear conv + Fully Connected &17.56\% \\
      % linear conv + Fully Connected + Dropout &15.99\% \\
      % linear conv + Global Average Pooling &16.46\% \\
      % \hline
    \end{tabular}
  \end{center}
\end{table}

As is shown in Table \ref{gapfc}, the fully connected layer without
dropout regularization gave the worst performance (11.59\%). This is
expected as the fully connected layer overfits to the training data if
no regularizer is applied. Adding dropout before the fully connected
layer reduced the testing error (10.88\%). Global average pooling has
achieved the lowest testing error (10.41\%) among the three.

We then explore whether the global average pooling has the same
regularization effect for conventional CNNs. We instantiate a
conventional CNN as described by Hinton et
al. \cite{hinton2012improving}, which consists of three convolutional
layers and one local connection layer. The local connection layer
generates 16 feature maps which are fed to a fully connected layer
with dropout. To make the comparison fair, we reduce the number of
feature map of the local connection layer from 16 to 10, since only
one feature map is allowed for each category in the global average
pooling scheme. An equivalent network with global average pooling is
then created by replacing the dropout + fully connected layer with
global average pooling. The performances were tested on the CIFAR-10
dataset.

This CNN model with fully connected layer can only achieve the error
rate of 17.56\%. When dropout is added we achieve a similar
performance (15.99\%) as reported by Hinton et
al. \cite{hinton2012improving}. By replacing the fully connected layer
with global average pooling in this model, we obtain the error rate of
16.46\%, which is one percent improvement compared with the CNN without
dropout. It again verifies the effectiveness of the global average
pooling layer as a regularizer.  Although it is slightly worse than
the dropout regularizer result, we argue that the global average
pooling might be too demanding for linear convolution layers as it
requires the linear filter with rectified activation to model the
confidence maps of the categories.

\subsection{Visualization of NIN}

We explicitly enforce feature maps in the last mlpconv layer of NIN to
be confidence maps of the categories by means of global average
pooling, which is possible only with stronger local receptive field
modeling, e.g. mlpconv in NIN. To understand how much this purpose is
accomplished, we extract and directly visualize the feature maps from
the last mlpconv layer of the trained model for CIFAR-10.

\begin{figure}[h]
  \centering
  \includegraphics[width=1.0\linewidth]{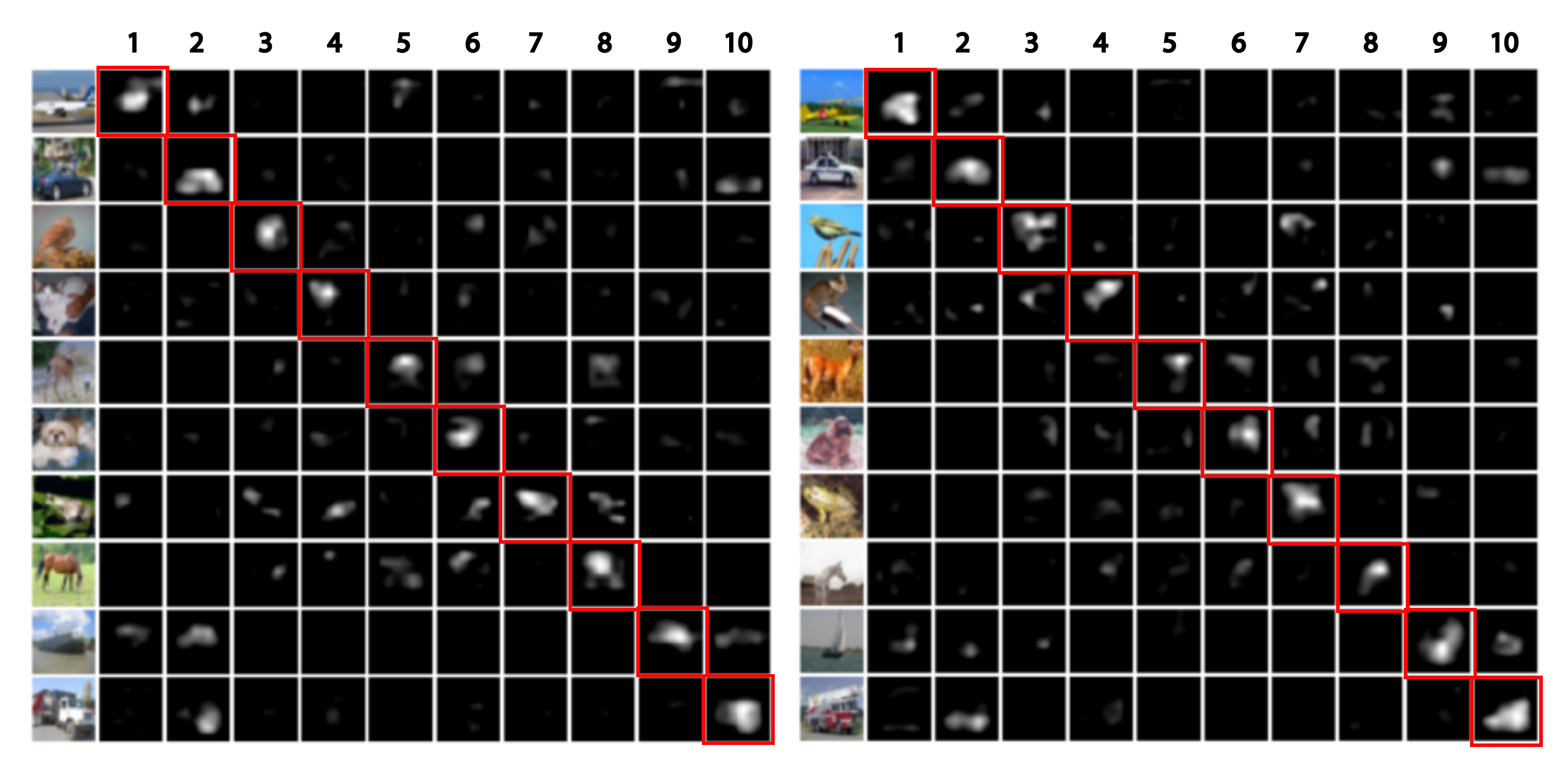}
  \caption{Visualization of the feature maps from the last mlpconv
    layer. Only top 10\% activations in the feature maps are
    shown. The categories corresponding to the feature maps are:
    1. airplane, 2. automobile, 3. bird, 4. cat, 5. deer, 6. dog,
    7. frog, 8. horse, 9. ship, 10. truck. Feature maps
    corresponding to the ground truth of the input images are
    highlighted. The left panel and right panel are just different
    examplars.}
  \label{fig:vis}
\end{figure}

Figure \ref{fig:vis} shows some examplar images and their
corresponding feature maps for each of the ten categories selected
from CIFAR-10 test set. It is expected that the largest activations
are observed in the feature map corresponding to the ground truth
category of the input image, which is explicitly enforced by global
average pooling. Within the feature map of the ground truth category,
it can be observed that the strongest activations appear roughly at
the same region of the object in the original image. It is especially
true for structured objects, such as the car in the second row of
Figure \ref{fig:vis}. Note that the feature maps for the categories
are trained with only category information. Better results are
expected if bounding boxes of the objects are used for fine grained
labels.

The visualization again demonstrates the effectiveness of NIN.  It is
achieved via a stronger local receptive field modeling using mlpconv
layers. The global average pooling then enforces the learning of
category level feature maps. Further exploration can be made towards
general object detection. Detection results can be achieved based on
the category level feature maps in the same flavor as in the scene
labeling work of Farabet et al. \cite{farabet2013learning}.

\section{Conclusions}

We proposed a novel deep network called ``Network In Network'' (NIN)
for classification tasks. This new structure consists of mlpconv
layers which use multilayer perceptrons to convolve the input and a
global average pooling layer as a replacement for the fully connected
layers in conventional CNN. Mlpconv layers model the local patches
better, and global average pooling acts as a structural regularizer
that prevents overfitting globally. With these two components of NIN
we demonstrated state-of-the-art performance on CIFAR-10, CIFAR-100
and SVHN datasets. Through visualization of the feature maps, we
demonstrated that feature maps from the last mlpconv layer of NIN were
confidence maps of the categories, and this motivates the possibility of
performing object detection via NIN.

\bibliographystyle{unsrt} \bibliography{iclr2014}

\begin{thebibliography}{10}

\bibitem{lecun1998gradient}
Yann LeCun, L{\'e}on Bottou, Yoshua Bengio, and Patrick Haffner.
\newblock Gradient-based learning applied to document recognition.
\newblock {\em Proceedings of the IEEE}, 86(11):2278--2324, 1998.

\bibitem{bengio2013representation}
Y~Bengio, A~Courville, and P~Vincent.
\newblock Representation learning: A review and new perspectives.
\newblock {\em IEEE transactions on pattern analysis and machine intelligence},
  35:1798--1828, 2013.

\bibitem{rosenblatt1961principles}
Frank Rosenblatt.
\newblock Principles of neurodynamics. perceptrons and the theory of brain
  mechanisms.
\newblock Technical report, DTIC Document, 1961.

\bibitem{krizhevsky2012imagenet}
Alex Krizhevsky, Ilya Sutskever, and Geoff Hinton.
\newblock Imagenet classification with deep convolutional neural networks.
\newblock In {\em Advances in Neural Information Processing Systems 25}, pages
  1106--1114, 2012.

\bibitem{hinton2012improving}
Geoffrey~E Hinton, Nitish Srivastava, Alex Krizhevsky, Ilya Sutskever, and
  Ruslan~R Salakhutdinov.
\newblock Improving neural networks by preventing co-adaptation of feature
  detectors.
\newblock {\em arXiv preprint arXiv:1207.0580}, 2012.

\bibitem{le2011ica}
Quoc~V Le, Alexandre Karpenko, Jiquan Ngiam, and Andrew Ng.
\newblock Ica with reconstruction cost for efficient overcomplete feature
  learning.
\newblock In {\em Advances in Neural Information Processing Systems}, pages
  1017--1025, 2011.

\bibitem{goodfellow2013piecewise}
Ian~J Goodfellow.
\newblock Piecewise linear multilayer perceptrons and dropout.
\newblock {\em arXiv preprint arXiv:1301.5088}, 2013.

\bibitem{goodfellow2013maxout}
Ian~J Goodfellow, David Warde-Farley, Mehdi Mirza, Aaron Courville, and Yoshua
  Bengio.
\newblock Maxout networks.
\newblock {\em arXiv preprint arXiv:1302.4389}, 2013.

\bibitem{gulccehre2013knowledge}
{\c{C}}a{\u{g}}lar G{\"u}l{\c{c}}ehre and Yoshua Bengio.
\newblock Knowledge matters: Importance of prior information for optimization.
\newblock {\em arXiv preprint arXiv:1301.4083}, 2013.

\bibitem{rowley1995human}
Henry~A Rowley, Shumeet Baluja, Takeo Kanade, et~al.
\newblock {\em Human face detection in visual scenes}.
\newblock School of Computer Science, Carnegie Mellon University Pittsburgh,
  PA, 1995.

\bibitem{zeiler2013stochastic}
Matthew~D Zeiler and Rob Fergus.
\newblock Stochastic pooling for regularization of deep convolutional neural
  networks.
\newblock {\em arXiv preprint arXiv:1301.3557}, 2013.

\bibitem{krizhevsky2009learning}
Alex Krizhevsky and Geoffrey Hinton.
\newblock Learning multiple layers of features from tiny images.
\newblock {\em Master's thesis, Department of Computer Science, University of
  Toronto}, 2009.

\bibitem{netzer2011reading}
Yuval Netzer, Tao Wang, Adam Coates, Alessandro Bissacco, Bo~Wu, and Andrew~Y
  Ng.
\newblock Reading digits in natural images with unsupervised feature learning.
\newblock In {\em NIPS Workshop on Deep Learning and Unsupervised Feature
  Learning}, volume 2011, 2011.

\bibitem{snoek2012practical}
Jasper Snoek, Hugo Larochelle, and Ryan~P Adams.
\newblock Practical bayesian optimization of machine learning algorithms.
\newblock {\em arXiv preprint arXiv:1206.2944}, 2012.

\bibitem{wan2013regularization}
Li~Wan, Matthew Zeiler, Sixin Zhang, Yann~L Cun, and Rob Fergus.
\newblock Regularization of neural networks using dropconnect.
\newblock In {\em Proceedings of the 30th International Conference on Machine
  Learning (ICML-13)}, pages 1058--1066, 2013.

\bibitem{malinowski2013learnable}
Mateusz Malinowski and Mario Fritz.
\newblock Learnable pooling regions for image classification.
\newblock {\em arXiv preprint arXiv:1301.3516}, 2013.

\bibitem{srivastava2013discriminative}
Nitish Srivastava and Ruslan Salakhutdinov.
\newblock Discriminative transfer learning with tree-based priors.
\newblock In {\em Advances in Neural Information Processing Systems}, pages
  2094--2102, 2013.

\bibitem{srivastava2013improving}
Nitish Srivastava.
\newblock {\em Improving neural networks with dropout}.
\newblock PhD thesis, University of Toronto, 2013.

\bibitem{goodfellow2013multi}
Ian~J Goodfellow, Yaroslav Bulatov, Julian Ibarz, Sacha Arnoud, and Vinay Shet.
\newblock Multi-digit number recognition from street view imagery using deep
  convolutional neural networks.
\newblock {\em arXiv preprint arXiv:1312.6082}, 2013.

\bibitem{farabet2013learning}
Cl{\'e}ment Farabet, Camille Couprie, Laurent Najman, Yann Lecun, et~al.
\newblock Learning hierarchical features for scene labeling.
\newblock {\em IEEE Transactions on Pattern Analysis and Machine Intelligence},
  35:1915--1929, 2013.

\end{thebibliography}

\end{document}

\maketitle

\section{Detailed settings for NIN on the datasets}

In this section, the detailed parameters and settings for running the
4 datasets are shown in tables. Since we adopted shared mode for all
the experiments, each layer of the mlp can be viewed as cascadable
cross channel parameteric pooling (cccp pooling). It is simpler and
more flexible to implement the mlpconv layer as stacked cccp layers.

\subsection{CIFAR-10}
\begin{table}[h!]
  \caption{Network structure for CIFAR-10}
  \label{cifar10}
  \begin{center}
    \begin{tabular}{l|l}
      \hline
      Layer            &Details \\
      \hline
      1. convolution   &kernel: 5, output: 192, pad: 2 \\
      2. cccp          &output: 160 \\
      3. cccp          &output: 96 \\
      4. pooling       &pool: max, kernel: 3, stride: 2 \\
      5. dropout       &ratio: 0.5 \\
      6. convolution   &kernel: 5, output: 192, pad: 2 \\
      7. cccp          &output: 192 \\
      8. cccp          &output: 192 \\
      9. pooling       &pool: max, kernel: 3, stride: 2 \\
      10. dropout      &ratio: 0.5 \\
      11. convolution  &kernel: 3, output: 192, pad: 1 \\
      12. cccp         &output: 192 \\
      13. cccp         &output: 10 \\
      14. gap          & \\
      \hline
    \end{tabular}
  \end{center}
\end{table}

\subsection{CIFAR-100}

CIFAR-100 shares exactly the same structure as CIFAR-10, the only
exception is the last cccp layer generates 100 channels, each channel
represents one category.

\subsection{SVHN}
\begin{table}[h!]
  \caption{Network structure for SVHN}
  \label{cifar10}
  \begin{center}
    \begin{tabular}{l|l}
      \hline
      Layer            &Details \\
      \hline
      1. convolution   &kernel: 5, output: 128, pad: 2 \\
      2. cccp          &output: 128 \\
      3. cccp          &output: 96 \\
      4. pooling       &pool: max, kernel: 3, stride: 2 \\
      5. dropout       &ratio: 0.5 \\
      6. convolution   &kernel: 5, output: 320, pad: 2 \\
      7. cccp          &output: 256 \\
      8. cccp          &output: 128 \\
      9. pooling       &pool: max, kernel: 3, stride: 2 \\
      10. dropout      &ratio: 0.5 \\
      11. convolution  &kernel: 5, output: 384, pad: 2 \\
      12. cccp         &output: 256 \\
      13. cccp         &output: 10 \\
      14. gap          & \\
      \hline
    \end{tabular}
  \end{center}
\end{table}

\subsection{MNIST}
\begin{table}[h!]
  \caption{Network structure for MNIST}
  \label{cifar10}
  \begin{center}
    \begin{tabular}{l|c}
      \hline
      Layer            &Details \\
      \hline
      1. convolution   &kernel: 5, output: 96, pad: 2 \\
      2. cccp          &output: 64 \\
      3. cccp          &output: 48 \\
      4. pooling       &pool: max, kernel: 3, stride: 2 \\
      5. dropout       &ratio: 0.5 \\
      6. convolution   &kernel: 5, output: 128, pad: 2 \\
      7. cccp          &output: 96 \\
      8. cccp          &output: 48 \\
      9. pooling       &pool: max, kernel: 3, stride: 2 \\
      10. dropout      &ratio: 0.5 \\
      11. convolution  &kernel: 5, output: 128, pad: 2 \\
      12. cccp         &output: 96 \\
      13. cccp         &output: 10 \\
      14. gap          & \\
      \hline
    \end{tabular}
  \end{center}
\end{table}

\section{Code}
The code to run NIN on the benchmark datasets and the instructions
have been made available on my github:
https://github.com/mavenlin/cuda-convnet